\documentclass[final,3p,times,twocolumn]{elsarticle}

\usepackage{amssymb}
\usepackage{amsmath}
\usepackage[table]{xcolor}
\usepackage{tabularx}
\usepackage{array}
\usepackage{multirow}
\usepackage{booktabs}
\usepackage[inline]{enumitem}  
\usepackage{bm}                
\usepackage{hyperref}

\journal{}

\makeatletter
\def\ps@pprintTitle{%
 \let\@oddhead\@empty
 \let\@evenhead\@empty
 \let\@oddfoot\@empty
 \let\@evenfoot\@empty
}
\makeatother

\makeatletter
\def\ps@pprintTitle{%
  \let\@oddhead\@empty
  \let\@evenhead\@empty
  \def\@oddfoot  {\reset@font\hfil\thepage\hfil}%
  \def\@evenfoot {\reset@font\hfil\thepage\hfil}%
}
\makeatother

\usepackage{fancyhdr}
\fancypagestyle{firstpage}{%
  \fancyhf{}                              
  \fancyfoot[L]{\footnotesize This paper is under consideration at \textit{Image and Vision Computing}.}
  \fancyfoot[R]{\footnotesize \today}
}

\begin{document}

\begin{frontmatter}





\title{MambaNeXt-YOLO: A Hybrid State Space Model for Real-time Object Detection}



\author[guet,gkl]{Xiaochun Lei\fnref{equalcontrib}}

\author[guet]{Siqi Wu\fnref{equalcontrib}\corref{cor1}}

\author[guet]{Weilin Wu}

\author[guet,gkl]{Zetao Jiang}

\cortext[cor1]{Corresponding author.}

\fntext[equalcontrib]{Equal contribution.}

\affiliation[guet]{
  organization={School of Computer Science and Information Security},
  addressline={Guilin University of Electronic Technology}, 
  city={Guilin},
  postcode={541010}, 
  state={Guangxi},
  country={China}
}

\affiliation[gkl]{
  organization={Guangxi Key Laboratory of Image and Graphic Intelligent Processing},
  addressline={Guilin University of Electronic Technology}, 
  city={Guilin},
  postcode={541010}, 
  state={Guangxi},
  country={China}
}

\begin{abstract}
Real-time object detection is a fundamental but challenging task in computer vision, particularly when computational resources are limited. Although YOLO-series models have set strong benchmarks by balancing speed and accuracy, the increasing need for richer global context modeling has led to the use of Transformer-based architectures. Nevertheless, Transformers have high computational complexity because of their self-attention mechanism, which limits their practicality for real-time and edge deployments. To overcome these challenges, recent developments in linear state space models, such as Mamba, provide a promising alternative by enabling efficient sequence modeling with linear complexity. Building on this insight, we propose MambaNeXt-YOLO, a novel object detection framework that balances accuracy and efficiency through three key contributions: (1) MambaNeXt Block: a hybrid design that integrates CNNs with Mamba to effectively capture both local features and long-range dependencies; (2) Multi-branch Asymmetric Fusion Pyramid Network (MAFPN): an enhanced feature pyramid architecture that improves multi-scale object detection across various object sizes; and (3) Edge-focused Efficiency: our method achieved 66.6\% mAP at 31.9 FPS on the PASCAL VOC dataset without any pre-training and supports deployment on edge devices such as the NVIDIA Jetson Xavier NX and Orin NX.
\end{abstract}




\end{frontmatter}
\thispagestyle{firstpage}

\section{Introduction}
\label{sec:intro}
Real-time object detection remains critical in computer vision, with diverse applications in autonomous driving, surveillance, robotics, and edge-computing scenarios such as UAV-assisted Internet of Vehicles, where deep reinforcement learning is used to optimize task offloading\cite{yan2023edge} and energy consumption\cite{yan2024energy}. In recent years, YOLO-series\cite{yolov1,yolov2,yolov3,yolov4,yolov5,yolov6,yolov7,yolov8,yolov9,yolov10,yolov12} models have become the preferred solutions for these scenarios due to their strong balance of speed and accuracy. Nonetheless, as real-world environments become increasingly complex, there is a growing need for models that efficiently capture both local features and global context while maintaining low computational overhead.

One promising direction is the Mamba\cite{mamba} architecture, a recent innovation based on state space models (SSMs)\cite{gu2021efficiently} that achieves linear time complexity for sequence modeling. Its efficiency and scalability have generated increasing interest in adapting Mamba for visual tasks. A notable example is MambaYOLO\cite{wang2024mambayolossmsbasedyolo}, which introduces a simple yet effective SSM-based backbone alongside a specialized RG Block to augment local feature modeling in object detection. This design sets a new baseline for real-time detection. However, although MambaYOLO demonstrates the potential of SSMs in object detection, it relies heavily on a monolithic architecture and does not explicitly explore the synergy between convolutional operations and SSMs, nor does it incorporate advanced multi-scale feature fusion strategies for detecting objects of varying sizes.

Motivated by these insights, we propose MambaNeXt-YOLO, a novel object detection framework that combines the local feature modeling strengths of CNNs with the efficient long-range sequence modeling capabilities of Mamba. Specifically, we introduce the MambaNeXt Block, a hybrid module designed to seamlessly integrate CNN layers and Mamba blocks for efficient global and local feature representation. To further improve multi-scale feature extraction, we adopted the Multi-branch Asymmetric Fusion Pyramid Network (MAFPN)\cite{yang2025mhaf}, which boosted detection performance across objects of varying sizes.

Our main contributions are as follows:
\begin{itemize}
    \item We introduce MambaNeXt, a novel hybrid block that combines CNNs with Mamba to capture both local features and long-range dependencies efficiently.
    \item To optimize multi-scale feature representation, we employed the MAFPN, which enhanced detection performance across objects of different sizes.
    \item Our method achieved 66.6\% mAP at 31.9 FPS on the PASCAL VOC dataset without any pre-training and was successfully deployed on resource-constrained edge devices such as the NVIDIA Jetson Xavier NX and Orin NX.
\end{itemize}

\section{Related Work}

\subsection{Real-time Object Detection}

Real-time object detection has long been a central focus in computer vision, with numerous models striving to strike a balance between accuracy and speed. The YOLO series has made significant advances by introducing single-stage detectors that directly predict bounding boxes and class probabilities. Later versions, such as YOLOv5\cite{yolov5} and YOLOv8\cite{yolov8}, incorporated more optimized backbones, lightweight modules, and improved feature fusion strategies, enhancing their suitability for edge applications. Nevertheless, their reliance on convolutional operations limits their capacity to capture global dependencies, which are essential for understanding complex scenes.

\subsection{Lightweight Vision Models and Hybrid Architectures}

In pure Transformer\cite{vaswani2017attention} architectures, the computation and memory requirements of self-attention scale quadratically as $\mathcal{O}(N^2)$ with the number of tokens $N$, creating a latency bottleneck for high-resolution inputs and resource-constrained devices. To overcome this inefficiency, various hybrid architectures have been developed. Models such as MobileViT\cite{mehta2021mobilevit}, EdgeViT\cite{pan2022edgevits}, and EfficientFormer\cite{li2022efficientformer} combine convolutional operations with lightweight attention or token-mixing mechanisms to strike a balance between speed and accuracy. These approaches use CNNs for efficient local processing while incorporating simplified attention blocks to capture long-range interactions. However, many still struggle to maintain competitive accuracy under strict latency constraints, particularly on low-power edge devices.

\subsection{State Space Models in Vision}

State space models (SSMs) have recently emerged as a promising alternative for sequence modeling, offering linear time complexity. Mamba introduced a selective state space mechanism that enables efficient modeling of long-range dependencies, attracting significant interest from both the NLP and vision communities. Visual adaptations such as Vision Mamba\cite{vim}, VMamba\cite{liu2024vmamba}, and LocalMamba\cite{huang2024localmamba} have demonstrated competitive performance in classification tasks by integrating SSMs with 2D visual structures.

In the field of object detection, Mamba YOLO is an early effort to integrate Mamba into a detection framework. It introduces the ODMamba backbone and RG Block to address the weak local modeling of SSMs. Nonetheless, Mamba YOLO relies mainly on a monolithic design and lacks explicit multi-scale fusion strategies, which limits its adaptability across different object scales.



\section{Method}

\begingroup
  \setlength{\textfloatsep}{0pt}  
  \setlength{\floatsep}{0pt}      
  \setlength{\intextsep}{0pt}     
  \begin{figure*}[!t]
    \centering
    \includegraphics[width=1\linewidth]{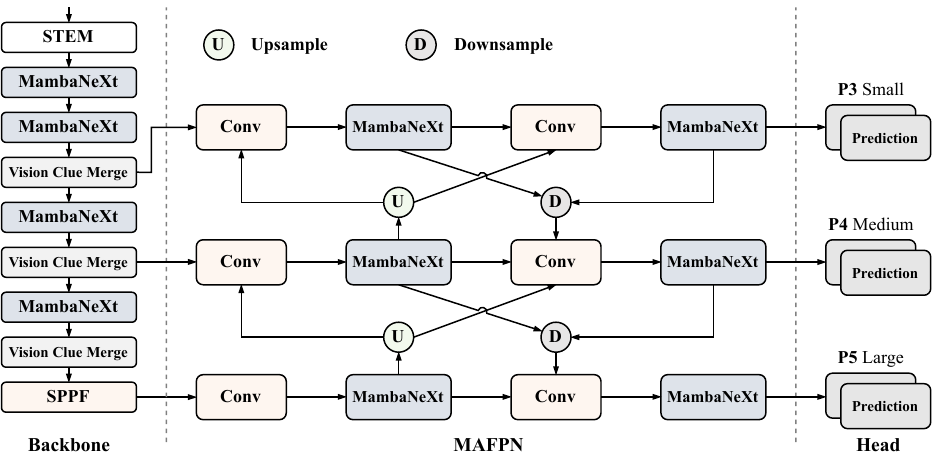}
    \caption{Illustration of the overall architecture of MambaNeXt-YOLO.}
    \label{fig:main}
  \end{figure*}
\endgroup

\begingroup
  \setlength{\textfloatsep}{0pt}  
  \setlength{\floatsep}{0pt}      
  \setlength{\intextsep}{0pt}     
  \begin{figure}[!t]
    \centering
    \includegraphics[width=1\linewidth]{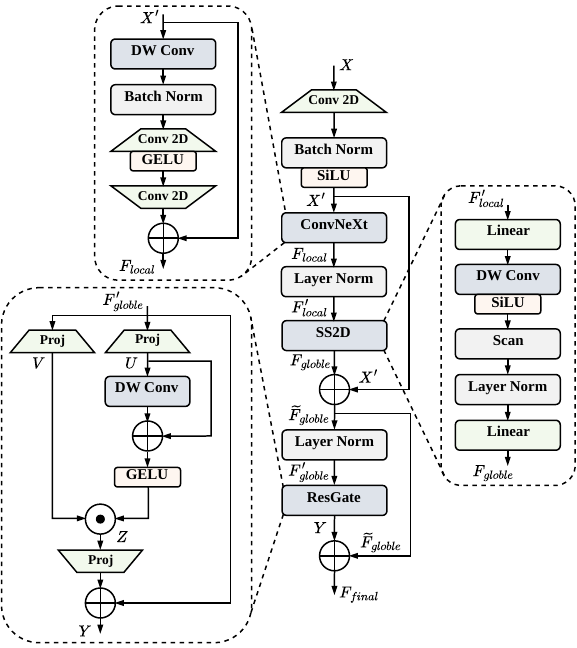}
    \caption{Illustration of the MambaNeXt block architecture.}
    \label{fig:block}
  \end{figure}
\endgroup

\begingroup
  \setlength{\textfloatsep}{0pt}  
  \setlength{\floatsep}{0pt}      
  \setlength{\intextsep}{0pt}     
  \begin{figure}[!t]
    \centering
    \includegraphics[width=1\linewidth]{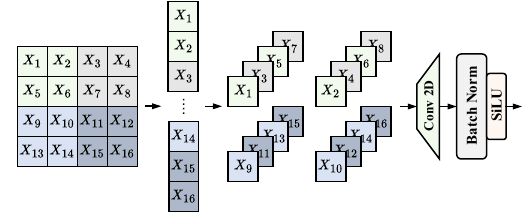}
    \caption{Illustration of the Vision Clue Merge Block.}
    \label{fig:vcm}
  \end{figure}
\endgroup

\subsection{Overview}
Given a color image $\mathbf{I} \in \mathbb{R}^{H \times W \times C}$, our goal is to predict a set of bounding boxes $\mathbf{b}_k$ with the corresponding class labels $y_k$. Here $H$, $W$, and $C$ denote the image height, width, and number of colour channels, respectively.

As sketched in Fig.~\ref{fig:main}, the proposed MambaNeXt-YOLO detector follows the classical \emph{backbone--neck--head} paradigm, with each component redesigned to optimize both accuracy and real-time efficiency. It consists of \begin {enumerate*}[label=(\roman*)]
\item a CNN--Mamba hybrid backbone that simultaneously captures local details and global context simultaneously,
\item a \textbf{M}ulti-\textbf{A}symmetric \textbf{F}usion \textbf{P}yramid \textbf{N}etwork (\textbf{MAFPN}) neck for cross-scale feature aggregation, and
\item a lightweight detection head that produces predictions at three resolutions (P3,\,P4,\,P5).
\end{enumerate*}

\paragraph{Backbone} The backbone begins with a shallow \emph{Stem} for patch embedding, followed by a cascade of \emph{MambaNeXt} blocks. Each block combines depthwise–separable convolutions with a Mamba state space model, allowing for the joint modelling of local spatial patterns and long-range dependencies. After each stage, a \emph{Vision Clue Merge} (VCM; Fig.~\ref{fig:vcm}) unit downsamples the feature map while preserving important features. Lastly, a \emph{Spatial Pyramid Pooling Fast} (SPPF) layer further expands the receptive field before passing the features to the neck.

\paragraph{Neck} Our redesigned MAFPN strengthens bidirectional information flow by integrating top-down and bottom-up pathways with asymmetric kernels, reparametrized convolutions, and Mamba modules. This design strengthens multi-scale feature representation while maintaining low latency.

\paragraph{Head} The fused features are finally decoded by a decoupled classification and regression head that processes three feature levels: P3 for small objects, P4 for medium objects, and P5 for large objects.

\subsection{Vision Clue Merge (VCM)}
To preserve informative visual cues during downsampling, we used the \emph{Vision Clue Merge} Block introduced in VMamba. Rather than applying a $3\times3$ convolution with stride 2, the feature map was first spatially split and then compressed using $1\times1$ pointwise convolutions. These compressed features are concatenated along the channel axis, and further reduced by another pointwise convolution, achieving a fourfold spatial reduction overall. Following Mamba-YOLO, all normalisation layers were removed from this unit to maintain a leaner implementation.

\subsection{MambaNeXt Block}

The MambaNeXt block (Fig.~\ref{fig:block}) consists of three parts: (a) a ConvNeXt block for local feature extraction, (b) an SS2D Mamba block for global context modeling, and (c) a ResGate block for adaptive fusion. $\odot$ denotes the Hadamard (element‐wise) product. Given an input feature map $\mathbf{X}\in\mathbb{R}^{H\times W\times C}$, we first apply a pointwise convolution followed by batch normalization and the SiLU activation to obtain the pre‐processed feature $\mathbf{X}'$:
\begin{equation}
\mathbf{X}' = \mathrm{SiLU}\bigl(\mathrm{BN}(\mathrm{Conv}_{1\times1}(\mathbf{X}))\bigr)
\end{equation}

\paragraph{(a) ConvNeXt local block} To capture fine‐grained spatial details, we applied a depthwise convolution on $\mathbf{X}'$, followed by batch normalization. A subsequent pointwise convolution expanded the channel dimension, which was then passed through a GELU activation and another pointwise convolution to generate the local feature map $\mathbf{F}_{\mathrm{local}}$:
\begin{align}
\mathbf{X}_{\mathrm{dw}}       &= \mathrm{DWConv}(\mathbf{X}')\\
\mathbf{X}_{\mathrm{bn}}       &= \mathrm{BN}(\mathbf{X}_{\mathrm{dw}})\\
\mathbf{F}_{\mathrm{local}}    &= \mathrm{Conv}_{1\times1}\bigl(\mathrm{GELU}(\mathrm{Conv}_{1\times1}(\mathbf{X}_{\mathrm{bn}}))\bigr)
\end{align}

\paragraph{(b) Mamba global block} The local feature \(\mathbf{F}_{\mathrm{local}}\) was first normalized using Layer Normalization (LN) to stabilize the feature distribution across channels:
\begin{equation}
\mathbf{F}'_{\mathrm{local}} = \mathrm{LN}(\mathbf{F}_{\mathrm{local}})
\end{equation}
It was then refined through a pointwise linear projection followed by an activation to refine the normalized features:
\begin{equation}
\mathbf{F}_{\mathrm{scan}} = \mathrm{SiLU}\bigl(\mathrm{DWConv}(\mathrm{Linear}(\mathbf{F}'_{\mathrm{local}}))\bigr)
\end{equation}
where \(\mathrm{Linear}(\cdot)\) is a \(1\times1\) convolution and \(\mathrm{DWConv}(\cdot)\) is a depthwise convolution. We then reshape \(\mathbf{F}_{\mathrm{scan}}\in\mathbb{R}^{C\times H\times W}\) into a sequence \(\{\mathbf{x}_t\}_{t=1}^{L}\) with \(L = H\times W\) by flattening the spatial dimensions.

At each time step $t$, the current token $\mathbf{x}_t$ was fed into three parallel linear projections to produce the input-conditioned parameters:
\begin{align}
\mathbf{A}_t &= \mathrm{Linear}_A(\mathbf{x}_t),\\
\mathbf{B}_t &= \mathrm{Linear}_B(\mathbf{x}_t),\\
\boldsymbol{\Delta}_t &= \mathrm{Linear}_{\Delta}(\mathbf{x}_t),
\end{align}
where $\mathrm{Linear}_A$, $\mathrm{Linear}_B$ and $\mathrm{Linear}_{\Delta}$ are learnable pointwise convolutions. The 1-D Mamba kernel then updated the hidden state $\mathbf{h}_t$ recursively:
\begin{equation}
\mathbf{h}_{t+1} = e^{-\boldsymbol{\Delta}_t}\odot \mathbf{h}_t \;+\; \mathbf{A}_t \;+\; \mathbf{B}_t\odot \mathbf{h}_t,
\end{equation}
with $\mathbf{h}_1$ initialized. After processing all $L$ tokens, the sequence of hidden states $\{\mathbf{h}_t\}_{t=1}^{L}$ was reshaped back into a feature map and further transformed to produce  the final global feature:
\begin{align}
\mathbf{F}_{\mathrm{mamba}}
&= \mathrm{reshape}\bigl(\{\mathbf{h}_t\}_{t=1}^{L}\bigr)
\quad\bigl(\in\mathbb{R}^{C\times H\times W}\bigr),\\
\mathbf{F}'_{\mathrm{mamba}}
&= \mathrm{LN}\bigl(\mathbf{F}_{\mathrm{mamba}}\bigr),\\
\mathbf{F}_{\mathrm{global}}
&= \mathrm{Linear}\bigl(\mathbf{F}'_{\mathrm{mamba}}\bigr).
\end{align}

\paragraph{(c) ResGate fusion block} Since \(\mathbf{F}_{\mathrm{global}}\) already encodes contextual information from \(\mathbf{F}_{\mathrm{local}}\), we first added a residual connection with the pre-processed input \(\mathbf{X}'\), followed by normalization:

\begin{align}
\widetilde{\mathbf{F}}_{global} &= \mathbf{F}_{\mathrm{global}} + \mathbf{X}' \\
\mathbf{F}'_{\mathrm{global}} &= \mathrm{LN}\bigl(\widetilde{\mathbf{F}}_{global}\bigr)
\end{align}

Next, two parallel pointwise projections generated gating activations \(U\) and \(V\):
\[
U = \mathrm{Proj}_2\bigl(\mathbf{F}'_{\mathrm{global}}\bigr),
\quad
V = \mathrm{Proj}_1\bigl(\mathbf{F}'_{\mathrm{global}}\bigr).
\]
A depthwise convolution combined with a GELU activation produced an information-bearing tensor, which was modulated by \(V\):
\[
Z = \mathrm{GELU}\bigl(\mathrm{DWConv}(U) + U\bigr)\;\odot\;V.
\]
Finally, a pointwise projection and second residual connection yield the fused output:
\[
\mathbf{Y} = \mathbf{F}'_{\mathrm{global}} + \mathrm{Proj}_3(Z).
\]

\paragraph{Final output}
The outputs of the global branch \(\mathbf{F}_{\mathrm{global}}\) and the fusion branch \(\mathbf{Y}\) are combined via a residual connection to form the final block output:
\begin{equation}
\mathbf{F}_{\mathrm{final}} = \widetilde{\mathbf{F}}_{global} + \mathbf{Y}
\end{equation}

\subsection{MAFPN}
We build upon the MAFPN from MHAF-YOLO, replacing its internal blocks with our MambaNeXt modules. In contrast to standard FPNs that use only a top-down pathway, MAFPN employs asymmetric branches to enable bidirectional information flow via both upsampling and downsampling. We substitute max-pooling with 3×3 convolutions of stride 2 followed by batch normalization, yielding more learnable and consistent feature transformations. At inference time, MambaNeXt’s multi-size kernels fuse into a single equivalent kernel, providing adaptive receptive fields without extra cost. As a result, our version achieves efficient multi-scale fusion, better small-object localization, and keeps the network lightweight.
\section{Experiment}
\subsection{Experimental Settings}
\paragraph{Dataset and evaluation} We evaluated our model on two benchmark datasets: Pascal VOC\cite{10.1007/s11263-009-0275-4} and DOTA v1.5\cite{Xia_2018_CVPR}. For Pascal VOC, the training set consisted of the combined train-val splits from VOC 2007 and VOC 2012, while the VOC 2007 test set was used for both validation and testing. To further assess the generalizability of our approach, we also included the DOTA v1.5 dataset in our evaluation. Performance was measured using the $AP_{\textbf{50}}$ and $AP_{\textbf{50:95}}$ metrics.

\paragraph{Implementation details} 

All experiments were conducted on a single NVIDIA RTX 3090 GPU using an input resolution of $640 \times 640$ and a batch size of 16. The model was trained for 1,000 epochs with the SGD optimizer, starting from an initial learning rate of 1e-2. For architecture-specific settings, the ConvNeXt blocks employed a layer scale initialization value of 1e-6 with default kernel configurations. The MambaNeXt module, including its core SS2D components, was configured with a state dimension of 16, a convolutional dimension of 3, and an SSM ratio factor of 2.0. As shown in Tables 1 and 2, MambaNeXt-YOLO consistently outperformed all baseline models across the mAP metrics on both the PASCAL VOC and DOTA v1.5 datasets. Specifically, it achieved 66.6\% mAP on PASCAL VOC and 27.8\% mAP on DOTA v1.5.


As shown in Table~\ref{tab:voc} and Table~\ref{tab:dota}, MambaNeXt-YOLO outperforms all baselines across mAP metrics in both PASCAL VOC and DOTAv1.5 dataset. MambaNeXt-YOLO achieves $66.6\%$ mAP on PASCAL VOC, and achieves $27.8\%$ mAP on DOTAv1.5.

\begin{table}[t]
\caption{Comparison of real-time object detectors in PASCAL VOC.}
\label{tab:voc}
\centering
\footnotesize
\setlength{\tabcolsep}{0.5em}
\begin{tabular}{c|c|c|c|c}
    \toprule
    Model & Params & FLOPs & $AP^{val}_{\textbf{50}}(\%)$ & $AP^{val}(\%)$ \\
    \midrule
    YOLOv8-S         & 11.1M & 28.7G & 83.9 & 64.4 \\
    YOLOv9-T         & 2.7M  & 11.1G & 84.1 & 65.5 \\
    YOLOv10-S        & 8.1M  & 24.9G & 84.3 & 65.0 \\
    YOLO11-S         & 9.4M  & 21.6G & 84.4 & 65.1 \\
    YOLOv12-S        & 9.1M  & 19.6G & 82.2 & 62.4 \\
    Gold-YOLO-N      & 5.8M  & 12.2G & 82.2 & 60.7 \\
    Hyper-YOLO-N     & 3.9M  & 11.0G & 82.6 & 62.9 \\
    Mamba-YOLO-T     & 5.9M  & 13.6G & 84.6 & 66.0 \\
    \rowcolor{gray!20}
    \textbf{MambaNeXt-YOLO} & 7.1M  & 22.4G & \textbf{85.1} & \textbf{66.6} \\
    \bottomrule
\end{tabular}
\end{table}

\begin{table}[t]
\caption{Comparison of real-time object detectors in DOTAv1.5.}
\label{tab:dota}
\centering
\footnotesize
\setlength{\tabcolsep}{0.5em}
\begin{tabular}{c|c|c|c|c}
    \toprule
    Model & Params & FLOPs & $AP^{val}_{\textbf{50}}(\%)$ & $AP^{val}(\%)$ \\
    \midrule
    YOLOv8-S         & 11.1M & 28.7G & 43.0 & 27.2 \\
    YOLOv9-T         & 2.7M  & 11.1G & 39.4 & 24.6 \\
    YOLOv10-S        & 8.1M  & 24.8G & 40.7 & 25.2 \\
    YOLO11-S         & 9.4M  & 21.6G & 43.5 & 27.6 \\
    YOLOv12-S        & 9.1M  & 19.6G & 43.1 & 27.7 \\
    Gold-YOLO-N      & 5.8M  & 12.2G & 13.0 & 5.3 \\
    Hyper-YOLO-N     & 3.9M  & 11.0G & 40.6 & 25.4 \\
    Mamba-YOLO-T     & 5.9M  & 13.6G & 41.3 & 25.7 \\
    \rowcolor{gray!20}
    \textbf{MambaNeXt-YOLO} & 7.1M  & 22.4G & \textbf{43.5} & \textbf{27.8} \\
    \bottomrule
\end{tabular}
\end{table}

\begin{table}[t]
\caption{Ablation study on MAFPN designs.}
\label{tab:ablation_mafpn}
\centering
\footnotesize
\setlength{\tabcolsep}{0.6em}
\begin{tabularx}{\linewidth}{>{\centering\arraybackslash}X|c|c|c|c}
    \toprule
    Method & Params & FLOPs & AP$^{\text{val}}_{\textbf{50}}$(\%) & AP$^{\text{val}}$(\%) \\
    \midrule
    Max-Pooling & 5.9M & 16.4G & 82.3 & 63.2 \\
    Conv        & 5.9M & 18.3G & \textbf{84.8} & \textbf{66.0} \\
    \bottomrule
\end{tabularx}
\end{table}

\begin{table}[t]
\caption{Ablation study on MambaNeXt block designs.}
\label{tab:ablation_mambanext}
\centering
\footnotesize
\setlength{\tabcolsep}{0.6em}
\begin{tabularx}{\linewidth}{>{\centering\arraybackslash}X|c|c|c|c}
    \toprule
    Method & Params & FLOPs & AP$^{\text{val}}_{\textbf{50}}$(\%) & AP$^{\text{val}}$(\%) \\
    \midrule
    ConvNeXt only & 6.3M & 15.2G & 82.3 & 63.3 \\
    ResGate only & 7.0M & 19.8G & 84.6 & 65.6 \\
    ConvNeXt first & 7.2M & 22.4G & 85.0 & 65.9 \\
    ResGate first & 7.1M & 22.4G & \textbf{85.1} & \textbf{66.2} \\
    \bottomrule
\end{tabularx}
\end{table}

\begin{table}[t]
\caption{Impact of Mamba module parameters on detection performance.}
\label{tab:ablation_mamba_params}
\centering
\footnotesize
\setlength{\tabcolsep}{0.5em}
\begin{tabularx}{\linewidth}{>{\centering\arraybackslash}X|c|c|c|c|c}
    \toprule
    Setting & Value & Params & FLOPs & AP$^{\text{val}}_{\textbf{50}}$(\%) & AP$^{\text{val}}$(\%) \\
    \midrule
    \multirow{3}{*}{$d_{state}$}
        & 8  & 6.9M & 22.4G & 85.1 & 65.9 \\
        & 16 & 7.1M & 22.4G & 85.1 & 66.2 \\
        & 32 & 7.7M & 22.4G & \textbf{85.1} & \textbf{66.6} \\
    \midrule
    \multirow{3}{*}{ssm\_ratio}
        & 1 & 6.2M & 20.4G & 84.1 & 64.9 \\
        & 2 & 7.1M & 22.4G & \textbf{85.1} & \textbf{66.2} \\
        & 4 & 11.3M & 36.6G & 84.3 & 65.4 \\
    \midrule
    \multirow{3}{*}{mlp\_ratio}
        & 2 & 6.4M & 19.8G & 85.0 & 65.8 \\
        & 4 & 7.1M & 22.4G & \textbf{85.1} & \textbf{66.2} \\
        & 8 & 8.7M & 27.7G & 84.4 & 65.7 \\
    \bottomrule
\end{tabularx}
\end{table}

\begin{table}[t]
\caption{Impact of ConvNeXt parameters on detection performance.}
\label{tab:ablation_convnext_params}
\centering
\footnotesize
\setlength{\tabcolsep}{0.6em}
\begin{tabularx}{\linewidth}{>{\centering\arraybackslash}X|c|c|c|c|c}
    \toprule
    Setting & Value & Params & FLOPs & AP$^{\text{val}}_{\textbf{50}}$(\%) & AP$^{\text{val}}$(\%) \\
    \midrule
    \multirow{3}{*}{layer\_scale}
        & $10^{-5}$ & 7.1M & 22.4G & 84.7 & 65.9 \\
        & $10^{-6}$ & 7.1M & 22.4G & \textbf{85.1} & \textbf{66.2} \\
        & $10^{-7}$ & 7.1M & 22.4G & 84.8 & 65.6 \\
    \midrule
    \multirow{3}{*}{dim} 
        & $2 \times $ & 6.4M & 19.9G & 84.4 & 65.3 \\
        & $4 \times $ & 7.1M & 22.4G & \textbf{85.1} & \textbf{66.2} \\
        & $8 \times $ & 8.7M & 27.5G & 84.8 & 66.1 \\
    \midrule
    \multirow{3}{*}{kernel\_size}
        & 3 & 7.1M & 22.2G & 84.4 & 65.1 \\
        & 5 & 7.1M & 22.3G & 84.7 & 65.7 \\
        & 7 & 7.1M & 22.4G & \textbf{85.1} & \textbf{66.2} \\
    \bottomrule
\end{tabularx}
\end{table}

\begin{table}[t]
\caption{Inference speed comparison on NVIDIA devices. Speed is measured in frames per second (FPS).}
\label{tab:inference_speed}
\centering
\footnotesize
\setlength{\tabcolsep}{0.55em}
\begin{tabularx}{\linewidth}{>{\centering\arraybackslash}X|c|c|c}
    \toprule
    Method & $\mathrm{FPS}_{\mathrm{Orin}}$ & $\mathrm{FPS}_{\mathrm{NX}}$ & $\mathrm{FPS}_{\mathrm{RTX3090}}$ \\
    \midrule
    YOLOv8-S         & 67.5 & 25.5  & 50.3 \\
    YOLOv9-T         & 27.1 & 18.7  & 36.8 \\
    YOLOv10-S        & 38.9 & 20.2  & 39.8 \\
    YOLO11-S         & 66.2 & 26.0  & 51.3 \\
    YOLOv12-S        & 36.5 & 19.0  & 37.5 \\
    Gold-YOLO-N      & 33.9 & 20.6  & 40.7 \\
    Hyper-YOLO-N     & 30.5 & 20.0  & 39.3 \\
    Mamba-YOLO-T     & 34.3 & 20.9  & 45.2 \\
    \cellcolor{gray!20}\textbf{MambaNeXt-YOLO} 
                     & \cellcolor{gray!20}31.9 
                     & \cellcolor{gray!20}19.5 
                     & \cellcolor{gray!20}34.6 \\
    \bottomrule
\end{tabularx}
\end{table}


\subsection{Ablation Studies on MAFPN Designs}
\paragraph{Impact of the downsampling method}

Table~\ref{tab:ablation_mafpn} compares different downsampling strategies within MAFPN, focusing on the effect of replacing conventional $2 \times 2$ max-pooling layers with stride convolutions (stride 2). The results indicated that convolution-based downsampling yielded a significant performance improvement, raising mAP from 63.2\% to 66.0\%.

\subsection{Ablation Studies on MambaNeXt Block Designs}
\paragraph{Impact of ConvNeXt and ResGate integration}

Table~\ref{tab:ablation_mambanext} presents an ablation study evaluating different configurations of the MambaNeXt block. Individually, both ConvNeXt and ResGate modules achieved solid performance, with ResGate offering a slight advantage due to its capacity for modeling adaptive feature relationships. However, combining both modules led to further gains. The configuration with ResGate preceding ConvNeXt (ResGate~$\rightarrow$~ConvNeXt) yielded the highest mAP at 66.2\%.

\paragraph{Impact of parameters of ConvNeXt}

Next, we also examined ConvNeXt-specific design parameters, including the layer scale initialization value, feature dimension width, and convolution kernel size. As shown in Table~\ref{tab:ablation_convnext_params}, setting the layer scale init value to 1e-6 delivered the best performance. Increasing the feature width to four times the input dimension resulted in optimal accuracy without introducing computational overhead. Furthermore, using a kernel size of 7 resulted in the highest mAP, confirming the benefit of larger receptive fields in the early stages of the network.

\paragraph{Impact of parameters of Mamba module}

We then investigated several key hyperparameters within the Mamba block, including the state dimension($d_{state}$), state space expansion ratio (SSM ratio), and MLP expansion ratio (MLP ratio). As shown in Table~\ref{tab:ablation_mamba_params}, increasing the $d_{state}$ from 8 to 32 steadily boosted mAP, peaking at 66.6\% when set to 32. The optimal SSM ratio was 2, striking a balance between performance and computational efficiency. For the MLP ratio, the best result was achieved at a value of 4; further increasing it added more parameters without improving accuracy.

\subsection{Inference Efficiency on NVIDIA Devices}

To assess the deployment potential of our model, we benchmarked its inference speed on several NVIDIA hardware platforms, including the desktop RTX 3090 GPU and edge-oriented devices such as the Jetson Xavier NX and Jetson Orin. For the edge devices, all tests were conducted using TensorRT with FP16 precision and a batch size of 1. Table~\ref{tab:inference_speed} further summarizes these results. Of note, MambaNeXt-YOLO achieved a 66.6\% mAP while delivering efficient inference, reaching 31.9 FPS on the Orin.
\section{Conclusion}
Herein, we propose MambaNeXt-YOLO, a novel hybrid object detection framework that merges the efficiency of CNNs with the global modeling strengths of state space models, specifically Mamba. By introducing the MambaNeXt block alongside the MAFPN, our approach effectively captures both fine-grained local details and long-range dependencies, achieving strong performance across multiple object scales. Extensive experiments on PASCAL VOC and DOTA v1.5 showed that MambaNeXt-YOLO achieved both higher accuracy and competitive real-time inference speeds, even on resource-limited devices like the Jetson Orin NX and Xavier NX. Importantly, our results highlight the effectiveness of combining SSM and CNN architectures for real-time vision tasks, particularly in edge deployment settings. Future work will focus on extending MambaNeXt-YOLO to video object detection, multi-modal perception, and semantic- and emotion-driven dual latent variable generation for dialogue systems\cite{yan2023semantic} to demonstrate the broad applicability of this hybrid framework across diverse deep generative tasks.

\bibliographystyle{splncs04}
\bibliography{bibliography.bib}

\end{document}